%% file: main.tex
\documentclass[10pt,twocolumn,letterpaper]{article}

\usepackage{wacv}
\usepackage{times}
\usepackage{epsfig}
\usepackage{graphicx}
\usepackage{amsmath}
\usepackage{amssymb}
\usepackage{booktabs}
\usepackage[dvipsnames]{xcolor}

\usepackage{multirow}
\usepackage{bbm}
\usepackage{subcaption}
\usepackage{bm}
\usepackage{enumitem}
\setlist{nosep}
\usepackage{pifont}%
\newcommand{\cmark}{{\color{ForestGreen} \ding{51}} }%
\newcommand{\xmark}{{\color{WildStrawberry} \ding{55}} }%
\newcommand{\figcaption}[2]{
\caption{\textbf{#1} #2}
}
\usepackage{xcolor}

\def\wacvPaperID{1363} %

\wacvfinalcopy %

\ifwacvfinal
\usepackage[pagebackref,breaklinks,colorlinks]{hyperref}
\else
\usepackage[pagebackref=true,breaklinks=true,colorlinks,bookmarks=false]{hyperref}
\fi

\usepackage[capitalize]{cleveref}
\crefname{section}{Sec.}{Secs.}
\Crefname{section}{Section}{Sections}
\Crefname{table}{Table}{Tables}
\crefname{table}{Tab.}{Tabs.}

\begin{document}

\title{Synthehicle: Multi-Vehicle Multi-Camera Tracking in Virtual Cities}

\author{ Fabian Herzog \quad Junpeng Chen \quad Torben Teepe  \quad Johannes Gilg \\[0.1cm]\quad Stefan H\"ormann \quad Gerhard Rigoll \\[0.2cm]
  Technical University of Munich, Germany\\[0.1cm]
  {\tt\small \href{https://fubel.github.io/synthehicle-dataset/}{fubel.github.io/synthehicle-dataset/}}
}

\maketitle

\begin{abstract}
   Smart City applications such as intelligent traffic routing or accident prevention rely on computer vision methods for exact vehicle localization and tracking.
    Due to the scarcity of accurately labeled data, detecting and tracking vehicles in 3D from multiple cameras proves challenging to explore. 
    We present a massive synthetic dataset for multiple vehicle tracking and segmentation in multiple overlapping and non-overlapping camera views. 
    Unlike existing datasets, which only provide tracking ground truth for 2D bounding boxes, our dataset additionally contains perfect labels for 3D bounding boxes in camera- and world coordinates, depth estimation, and instance, semantic and panoptic segmentation. 
    The dataset consists of 17 hours of labeled video material, recorded from 340 cameras in 64 diverse day, rain, dawn, and night scenes, making it the most extensive dataset for multi-target multi-camera tracking so far.
    We provide baselines for detection, vehicle re-identification, and single- and multi-camera tracking. Code and data are publicly available.\footnote{Code and data: \href{https://github.com/fubel/synthehicle}{https://github.com/fubel/synthehicle}}
\end{abstract}

\input{content/01_introduction}

\input{content/02_related_work}

\input{content/03_dataset}
\input{content/04_experiments}

\input{content/05_conclusion}

{\small
\bibliographystyle{ieee_fullname}
\bibliography{output}
}

\end{document}

%% file: content/01_introduction.tex
\section{Introduction}

As cities grow larger in population, increased car traffic causes jams, pollution, and accidents. 
Future smart cities, where multiple sensors (e.g., RGB cameras) are placed near crossroads, could reduce these problems through intelligent traffic management driven by computer vision. 

In particular, multi-target multi-camera tracking (MTMCT) is an essential task for such methods as it enables 3D localization and provides information for scene understanding.
In MTMCT, distinct objects must be unambiguously tracked across multiple cameras through space and time. 
The tracking problem is generally challenging, even in the single-camera case, primarily due to occlusions. 
Information added by multiple cameras can be beneficial but complicates data processing.
Existing solutions for MTMCT \cite{DBLP:conf/cvpr/RistaniT18, he2020multi, qian2020electricity, luna2021online, DBLP:journals/corr/abs-2204-10380} are mainly based on tracking-by-detection, in which an object detector is applied to obtain frame-wise sets of 2D bounding boxes which are then processed by a deep neural feature extractor for data association. 
Recently, the single-camera tracking community has started paying more attention to more complex tasks, such as 3D multi-object tracking \cite{DBLP:conf/iros/WengWHK20, DBLP:conf/cvpr/YinZK21} and tracking and segmentation \cite{DBLP:conf/cvpr/VoigtlaenderKOL19, DBLP:conf/nips/WeberXCZVCG0LCO21}. 
Data annotation, particularly for multi-camera setups and 3D localization and segmentation, is time-consuming and expensive. To this day, there is no real dataset for multi-target multi-camera tracking with 3D and segmentation annotations. Both enable potentially better 3D localization than approaches based on 2D bounding boxes.
\begin{figure}%
    \centering
    \subfloat[\centering 3D Boxes]{{\includegraphics[width=0.45\linewidth]{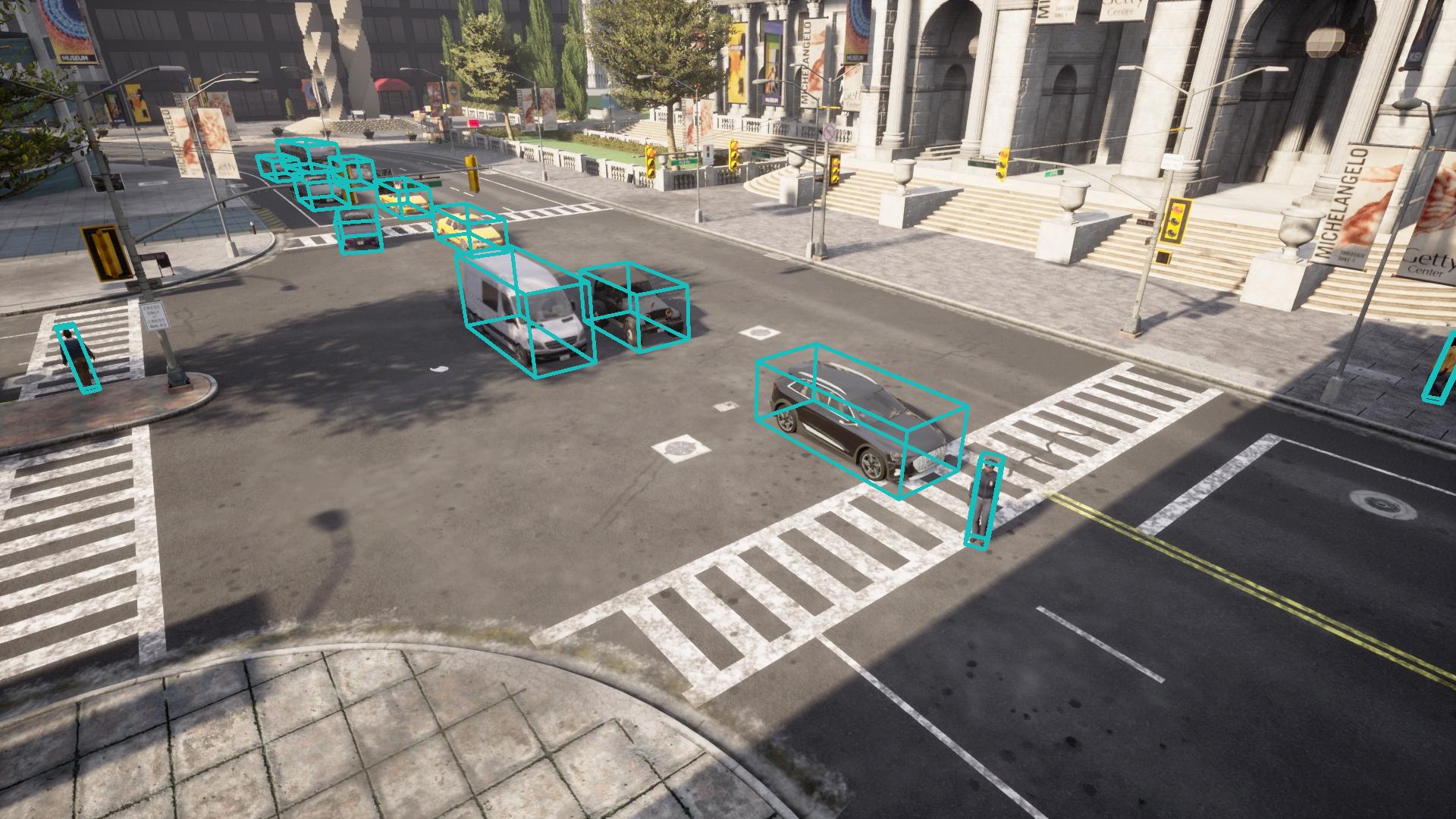} }}%
    \hfill
    \subfloat[\centering Semantic Segmentation]{{\includegraphics[width=0.45\linewidth]{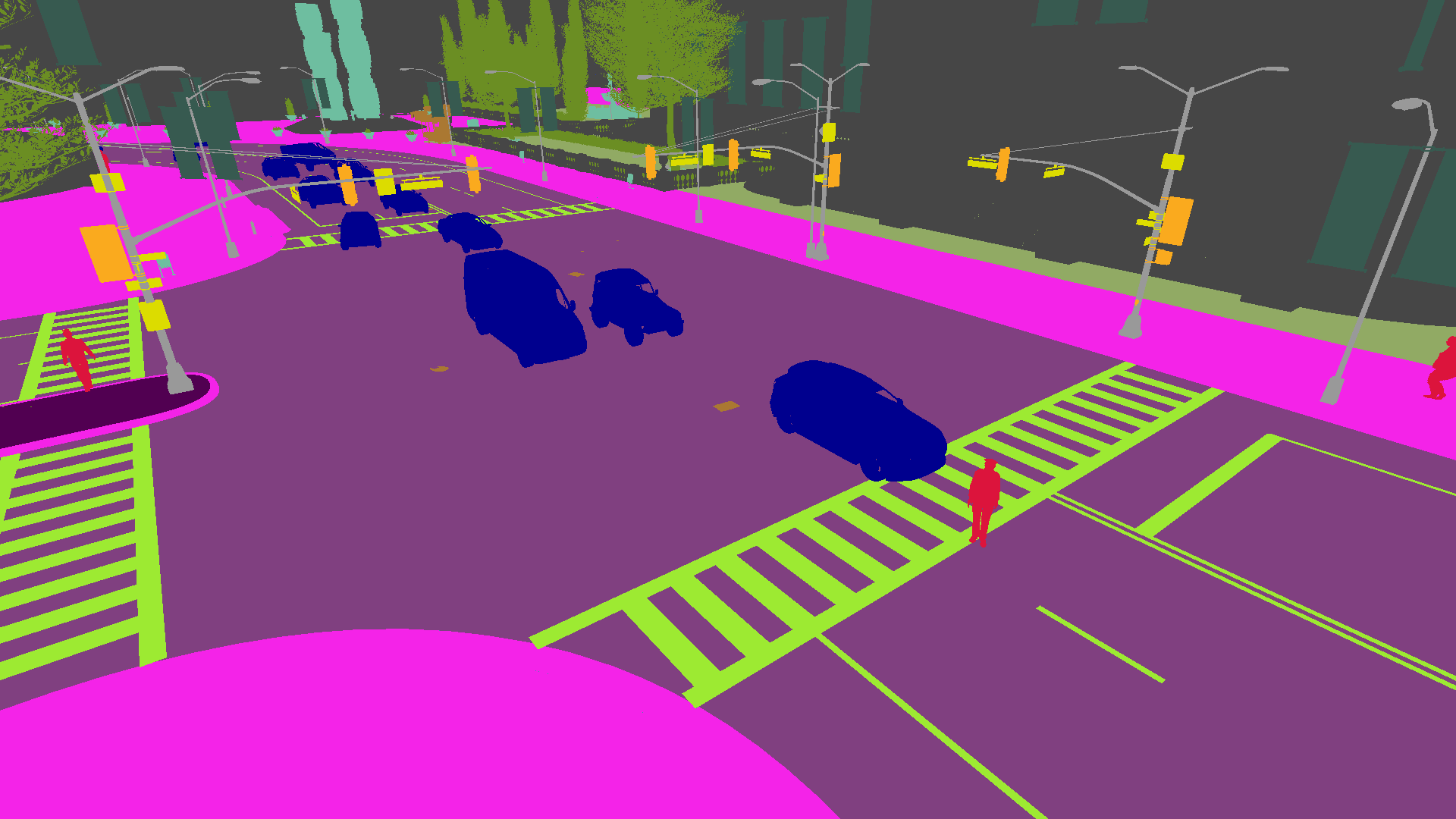} }}\\ \vspace{0.2cm}
    \subfloat[\centering Instance Segmentation]{{\includegraphics[width=0.45\linewidth]{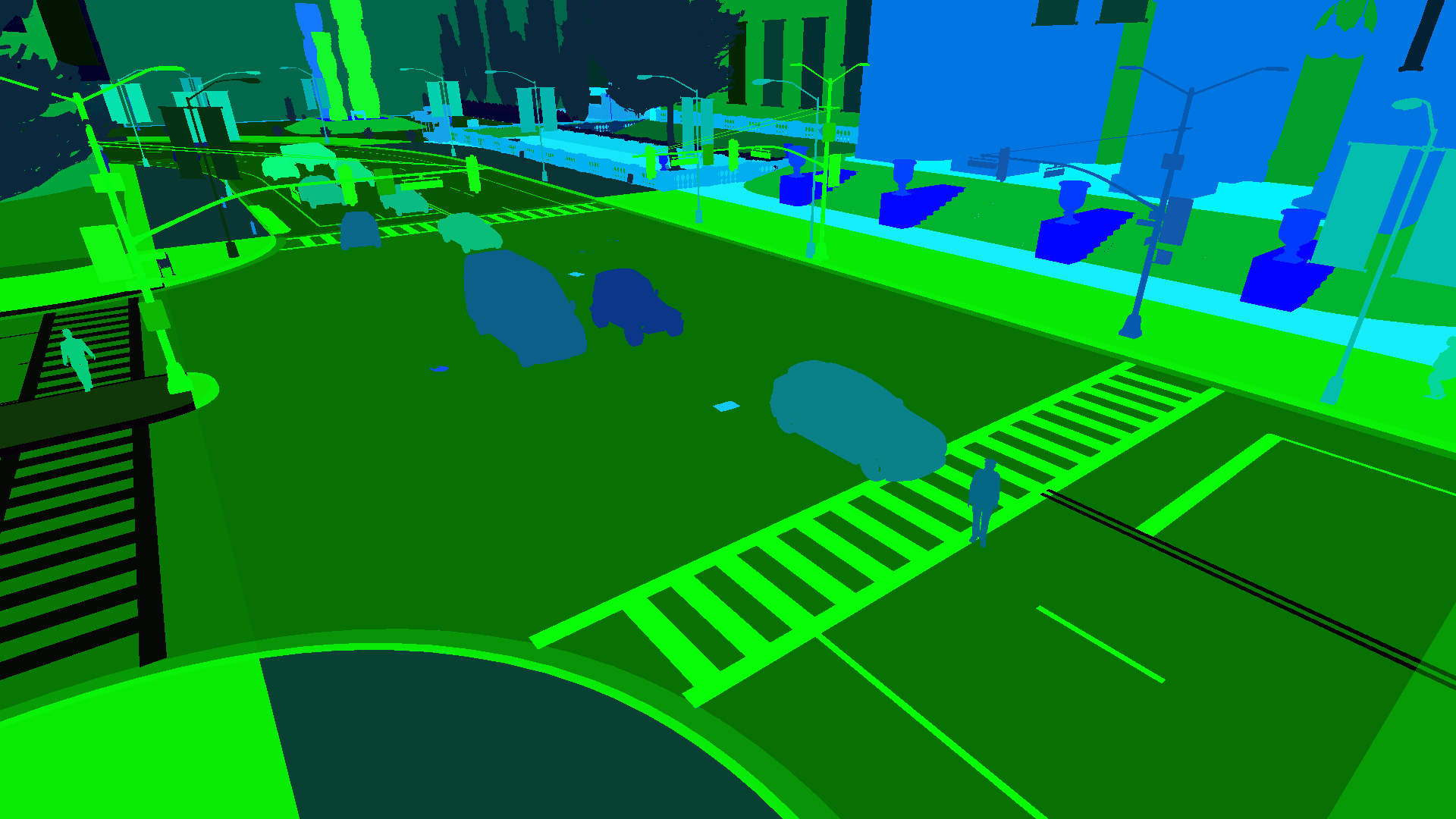} }}%
    \hfill
    \subfloat[\centering Depth Maps]{{\includegraphics[width=0.45\linewidth]{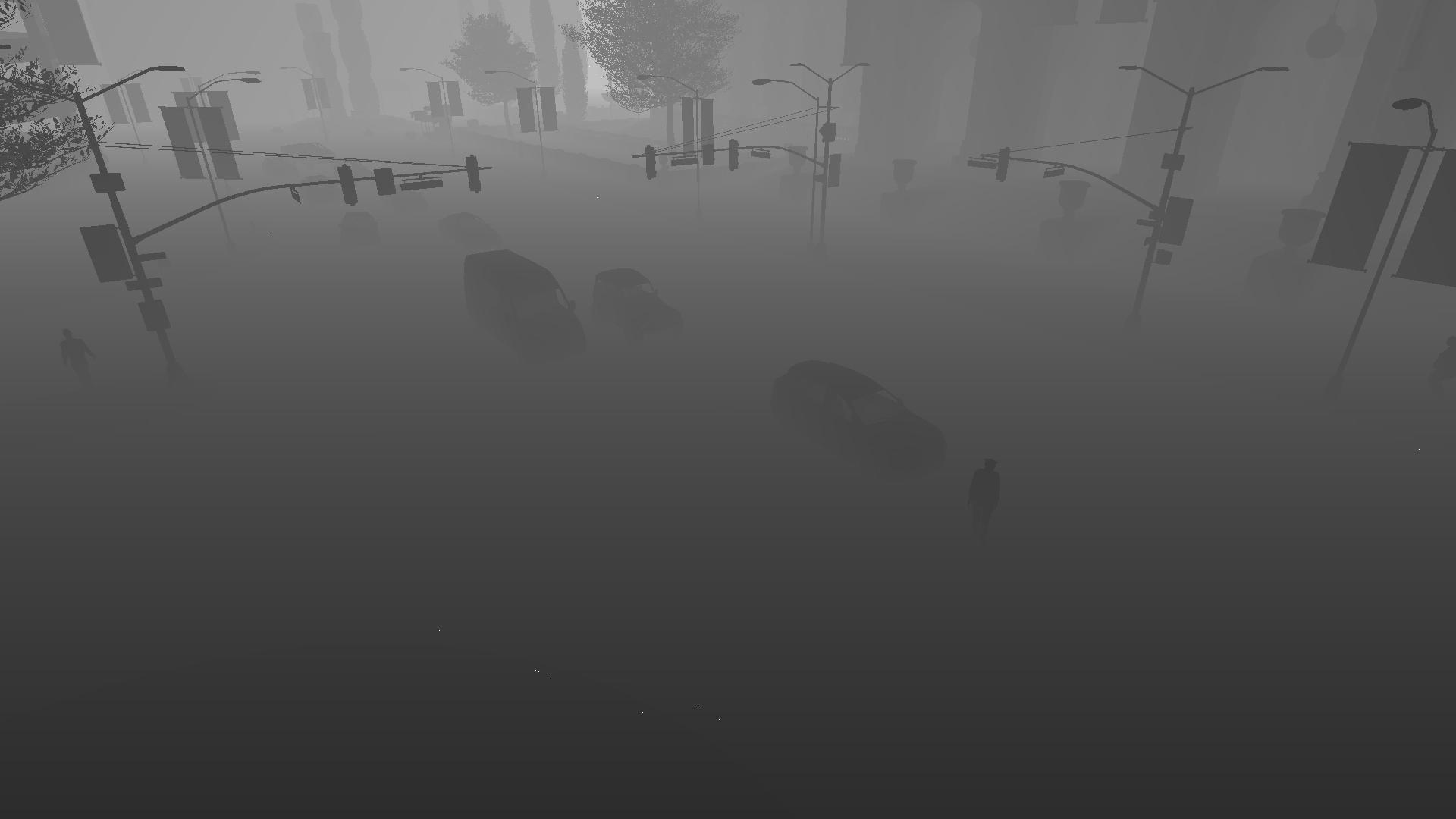} }}
    \figcaption{Ground truth annotation in Synthehicle.}{The proposed dataset contains perfect ground truth annotations for 3D detection (a), semantic (b), instance and panoptic (c) segmentation and depth estimation (d).}%
    \label{fig:example}%
\end{figure}

We address the scarcity of 3D and segmentation data and introduce \textit{Synthehicle}, a massive synthetic dataset for multiple vehicle detection, tracking, and segmentation across multiple cameras with overlapping and non-overlapping field of views (see Figure~\ref{fig:example} for example annotations).
The dataset consists of 17 hours of labeled video material, recorded from 340 cameras placed around crossroads and highways in 64 diverse day, rain, dawn and night scenes, and has been created using CARLA~\cite{Dosovitskiy17}. Summarized, our contributions are:

\begin{itemize}
    \item a massive synthetic dataset for multi-target multi-camera tracking with 2D, 3D, segmentation and depth annotations;
    \item a public evaluation server to test methods against ground truth of the test sets; and
    \item baseline results for 2D detection, vehicle re-identification, single- and multi-camera multi-vehicle tracking tasks.
\end{itemize}

Code for data generation, detection, re-identification and tracking, as well as all the generated data and evaluation scripts are publicly available.

%% file: content/02_related_work.tex
\section{Related Work}

\begin{figure*}%
    \centering
    \subfloat[\centering Day ]{{\includegraphics[width=0.23\linewidth]{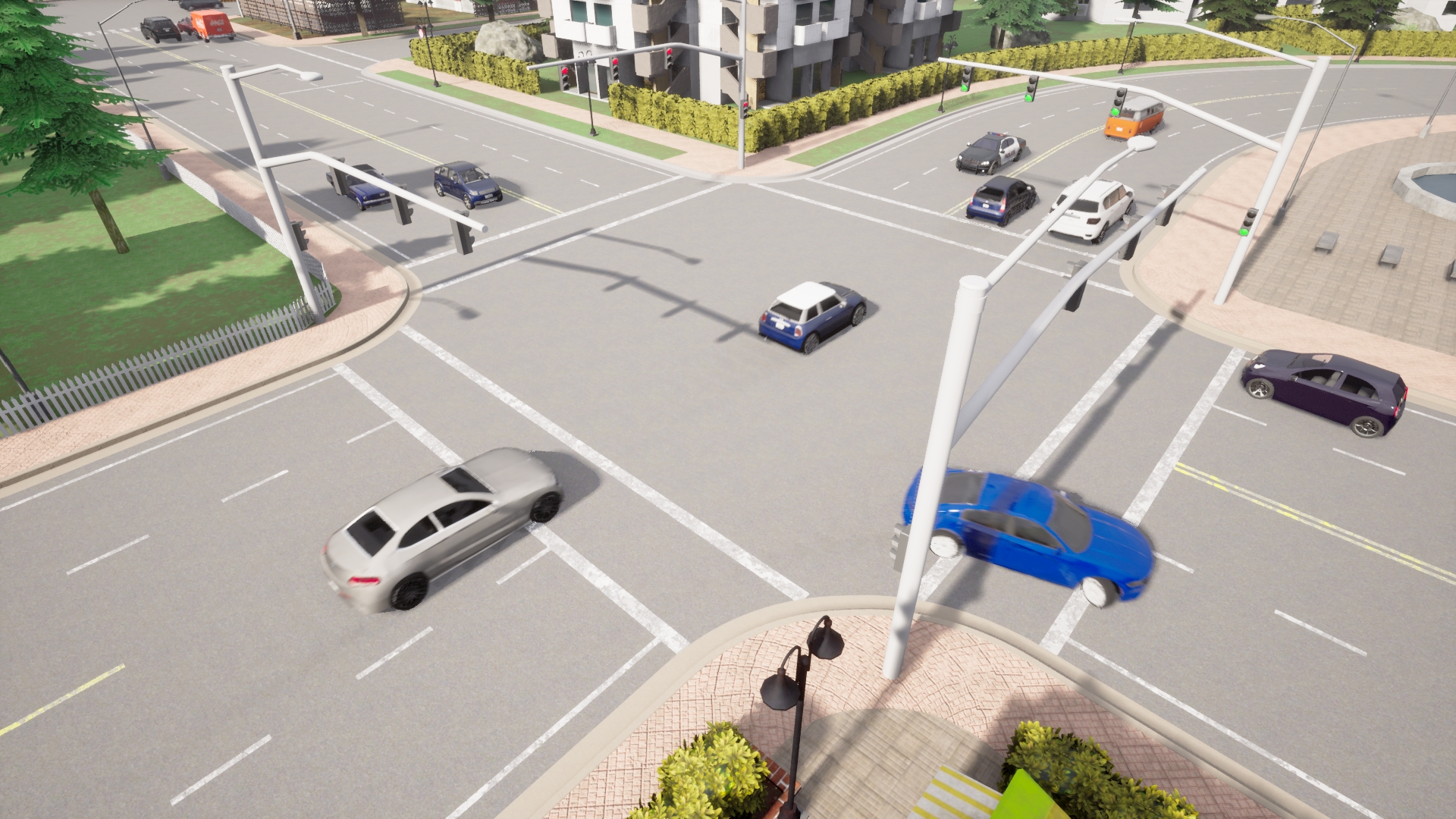} }}%
    \hfill
    \subfloat[\centering Dawn]{{\includegraphics[width=0.23\linewidth]{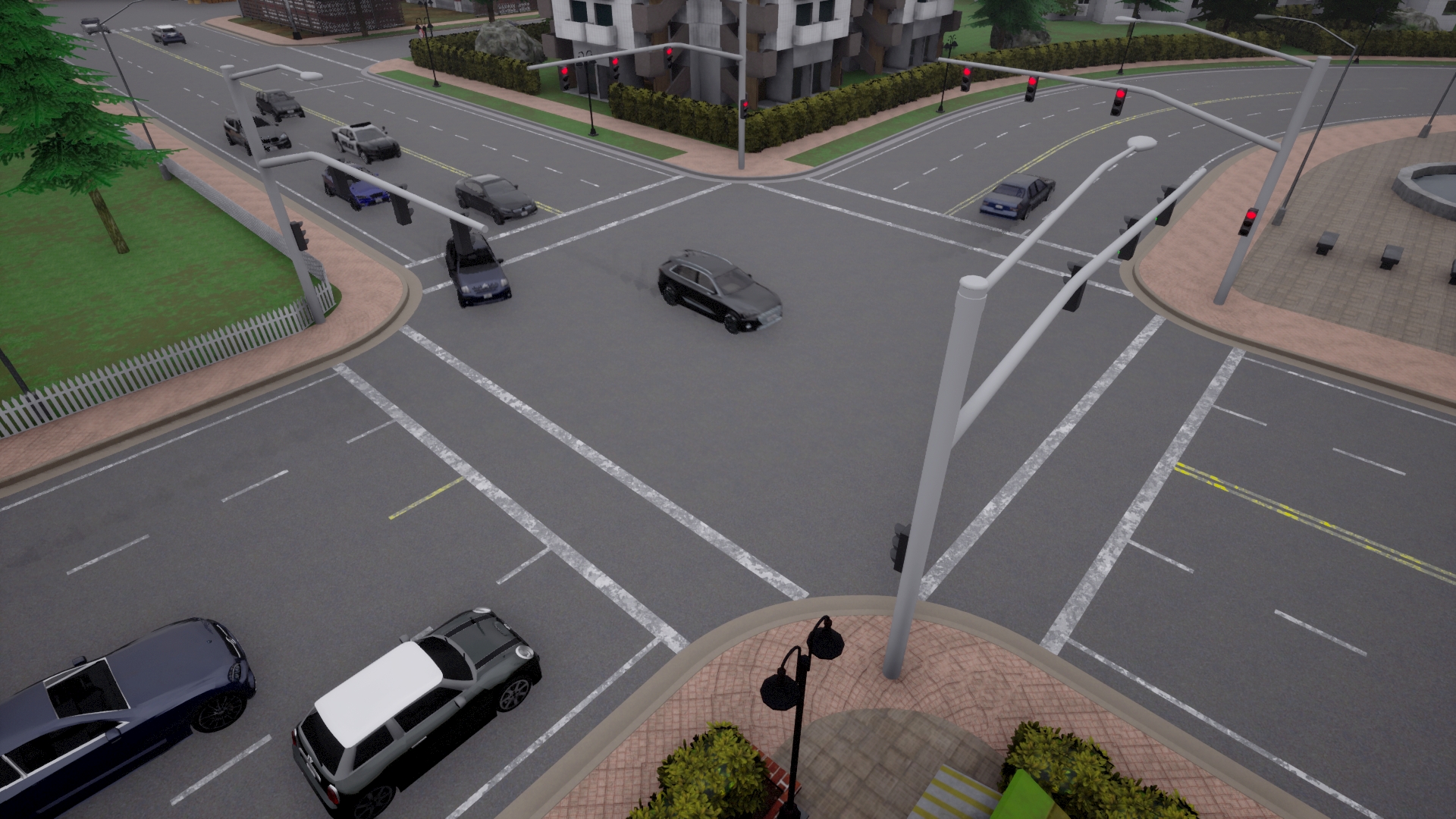} }}
    \hfill
    \subfloat[\centering Rain]{{\includegraphics[width=0.23\linewidth]{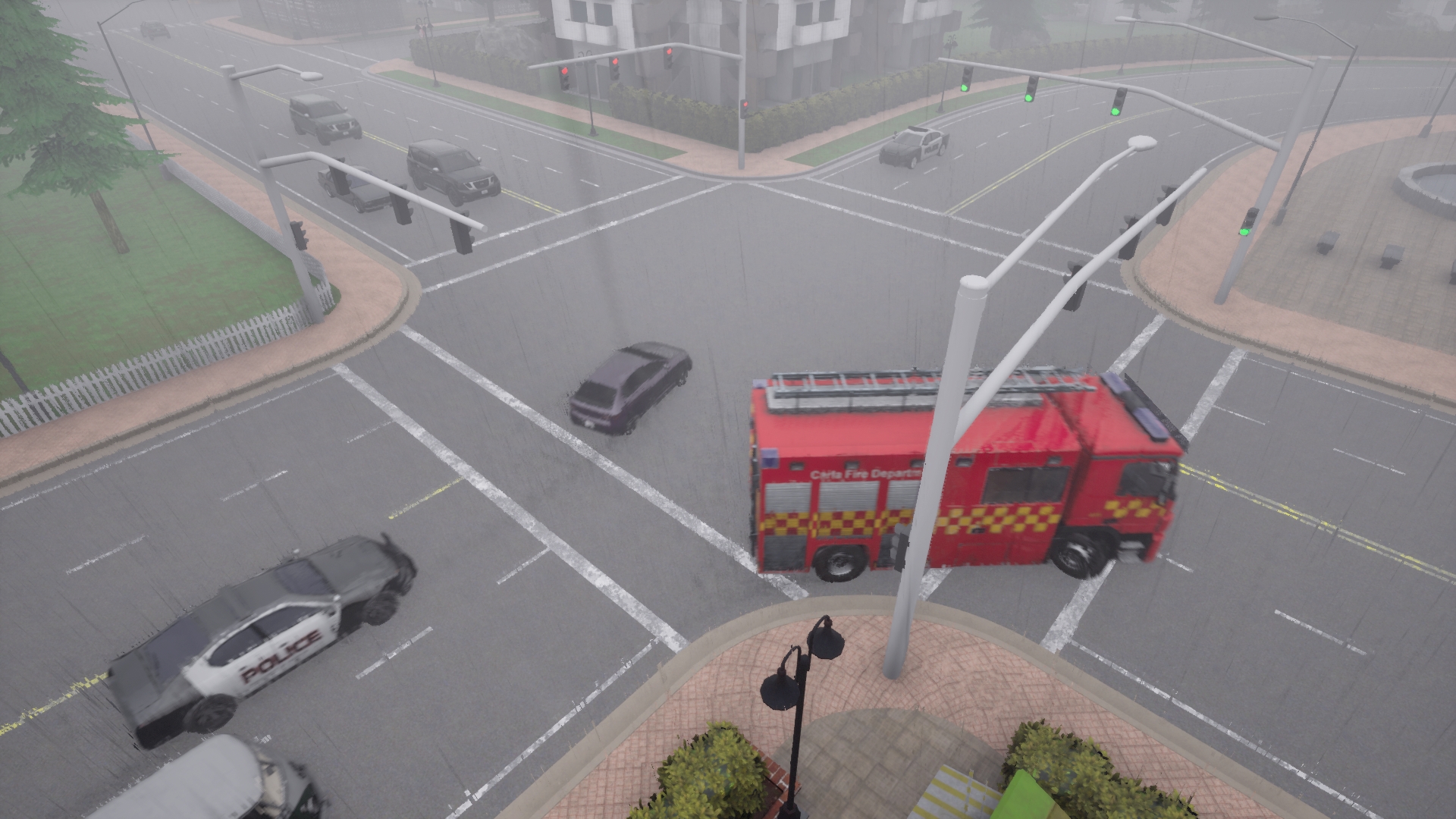} }}%
    \hfill
    \subfloat[\centering Night]{{\includegraphics[width=0.23\linewidth]{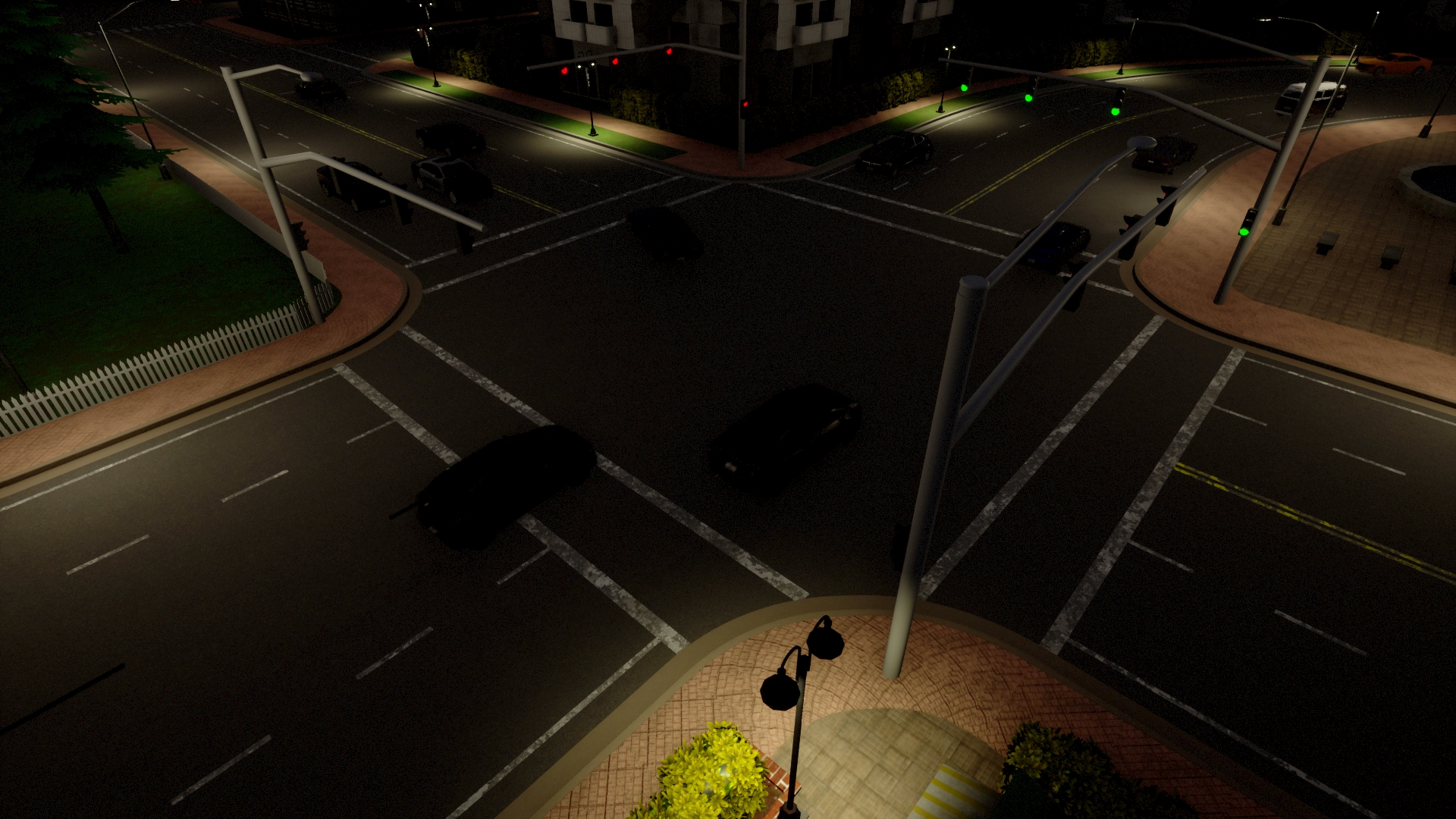} }}
    \figcaption{Ambience configurations.}{In Synthehicle, scenes are rendered under four different ambience configurations: day (a), dawn (b), rain (c), and night (d).}%
    \label{fig:weatherman}%
\end{figure*}
\paragraph{Vehicle Detection and Tracking Datasets} CityFlow~\cite{tang2019cityflow, Naphade20AIC20} is the established dataset to train and test vision methods for smart city applications such as multi-camera vehicle tracking, vehicle re-identification, and vehicle counting. However, it does not include 3D box annotations, depth maps, or segmentations.
KITTI~\cite{geiger2012kitti, geiger2013vision} is loosely related as it includes data for 2D and 3D detection and tracking of vehicles. As KITTI aims at improving autonomous vehicles, the data is captured from specific ego-motion camera angles. 
nuScenes~\cite{caesar2020nuscenes}, also designed for autonomous driving, is similar to KITTI, but significantly larger. It provides more ego-motion camera angles and includes night and rain scenarios.
Waymo Open \cite{DBLP:conf/cvpr/SunKDCPTGZCCVHN20} is similar in design but even more extensive than nuScenes. Neither KITTI, nuScenes nor Waymo Open provide the data for smart city scenarios. They focus on ego-motion and stereo vision by design and are unsuitable for studying problems such as vehicle re-identification and multi-vehicle multi-camera tracking.

\paragraph{Person Tracking Datasets} Many datasets have been proposed for the related tasks of single- and multi-camera \textit{person} tracking. For single-camera scenarios, the MOT challenges~\cite{MOT16, MOTChallenge20, MOTS20} are the established benchmarks to compare person trackers. The PETS09~\cite{ferryman2009pets2009} dataset can be considered the first relevant dataset for multi-camera person tracking. Other and larger datasets in this area are the EPFL-RLC~\cite{chavdarova2017epfl}, CAMPUS~\cite{xu2016multi}, MCT~\cite{chen2014mct} datasets. WILDTRACK~\cite{chavdarova2018wildtrack} is an extensive HD dataset developed for multiview detection and tracking, among other things. The DukeMTMC dataset, the largest real multi-camera tracking dataset, is no longer available due to privacy issues \cite{ristani2016performance, harvey2019megapixels}. To this day, there is no real dataset to substitute it.
Recent trends in single-camera person tracking are based on 3D detections \cite{DBLP:conf/iros/WengWHK20, DBLP:conf/cvpr/YinZK21} and tracking and segmentation \cite{DBLP:conf/cvpr/VoigtlaenderKOL19, DBLP:conf/nips/WeberXCZVCG0LCO21}, where the objects of interest have to be tracked pixel-wise. Both tasks allow a more exact localization of objects. Synthehicle is the first dataset to include 3D and segmentation ground truth for multi-camera multi-vehicle tracking.

\input{content/tables/dataset_comparison}

\paragraph{Vehicle Re-Identification Datasets} Just like person tracking relies on person feature extraction, most multi-vehicle trackers build upon vehicle re-identification methods~\cite{qian2020electricity, luna2021online, tran2022robust, specker2022improving, yao2022city, li2022multi, yang2022box}, which are usually trained on the VeRi-776 \cite{DBLP:conf/eccv/LiuLMM16, DBLP:conf/icmcs/LiuLMF16, DBLP:journals/tmm/LiuLMM18} or CityFlow re-identification \cite{tang2019cityflow} datasets. Compared to VeRi-776, CityFlow was recorded in more diverse scenarios and viewing angles. Because trackers have to deal with occlusions and bad lighting conditions, the VERI-Wild vehicle re-identification dataset~\cite{DBLP:conf/cvpr/LouB0WD19} has been proposed to provide images and annotations recorded in the wild.

\paragraph{Synthetic Datasets} Recording and labeling real data is time-consuming and expensive. In addition, data protection rights are potentially infringed when recording humans. Therefore, numerous synthetic datasets have been published to solve common computer vision tasks, such as segmentation \cite{DBLP:conf/cvpr/Varol0MMBLS17, DBLP:journals/corr/HandaPBSC15a, DBLP:conf/cvpr/RosSMVL16, DBLP:conf/cvpr/HuCHHS19, DBLP:conf/eccv/RichterVRK16, DBLP:conf/cvpr/Krahenbuhl18}, detection \cite{DBLP:conf/cvpr/MarinVGL10, DBLP:conf/iciap/AmatoCFGM19}, pose estimation \cite{fabbri2018learning}, and tracking \cite{fabbri2018learning, kohl2020mta, fabbri2021motsynth, DBLP:journals/corr/GaidonWCV16, deschaud2021kitticarla, DBLP:conf/iccv/HuCWLSKDY19}.
In person tracking, the most significant synthetic datasets are JTA~\cite{fabbri2018learning}, which presents a dataset for single-camera pose-tracking based on the video game GTA V. Closely related, MTA~\cite{kohl2020mta} offers multi-camera multi-person tracking data that can be seen as a replacement for the DukeMTMC dataset. MOTSynth showed that training on synthetic data can improve tracking results on real data~\cite{fabbri2021motsynth}. The synthetic VehicleX~\cite{Yao20VehicleX} dataset was generated for vehicle re-identification. Related datasets also based on CARLA are KITTI-CARLA~\cite{deschaud2021kitticarla}, which provides synthetic lidar data analogous to KITTI~\cite{geiger2012kitti}, V2I-CARLA~\cite{wang2022v2icarla} for synthetic vehicle re-identification, and the Paris-CARLA-3D~\cite{deschaud2021pariscarla} dataset for 3D mapping.

%% file: content/tables/dataset_comparison.tex
\begin{table*}[!ht]
\centering
 \resizebox{\textwidth}{!}{
\begin{tabular}{llrrcrrccccc}
\toprule
& \textbf{Dataset} & \textbf{\# Cams} & \textbf{\# Boxes} & \textbf{\# Scenes} & \textbf{Density} & \textbf{Duration} & \textbf{Targets}  & \textbf{3D Boxes}  & \textbf{Depth} & \textbf{Segmentation} \\ \midrule 
\multirow{6}{*}{\rotatebox[origin=c]{90}{Real}} & DukeMTMC (offline)~\cite{ristani2016DukeMTMC}& 8 & 4\,077\,132 & 1 & 1.9 & 11.33h & Person  & \xmark & \xmark & \xmark \\
&CAMPUS~\cite{xu2016multi} & 16 & 12\,264& 4 & -- & ~45m & Person   & \xmark & \xmark & \xmark   \\
&EPFL-RLC & 3&  6,132 & 1 & -- & 6.6 min  & Person  & \xmark & \xmark & \xmark  \\
&PETS09 & 7 & 4650 & 8 & --  & 2 min & Person   & \xmark & \xmark & \xmark  \\
&WILDTRACK~\cite{chavdarova2018wildtrack} & 7 & 56\,000 & 1 & 23.8 & 1h & Person  & \xmark & \xmark & \xmark\\
&CityFlow~\cite{tang2019cityflow} & 46 & 313\,931 & 6 & ~2.5 & 3.58h & Vehicles  & \xmark & \xmark & \xmark \\
\midrule
\multirow{4}{*}{\rotatebox[origin=c]{90}{Synthetic}}  
& JTA~\cite{fabbri2018learning} & 1* & 10\,000\,000& 512  & 20 & 4.27h & Person & \xmark & \xmark & \xmark \\
& MOTSynth~\cite{fabbri2021motsynth} & 1* & 40\,780\,800 & 768 & 29.5 & 15.36h & Person  & \xmark & \cmark & \cmark \\
& MTA~\cite{kohl2020mta} & 6 & 37\,324\,348 & 1 & 24.8 & 10.2h & Person  & \xmark & \xmark & \xmark \\
& \textbf{Synthehicle (ours)} & \textbf{340} & 4\,623\,184 & 64 & 7.45 & \textbf{17.00h} & Vehicles  & \cmark & \cmark & \cmark \\
\bottomrule
\end{tabular}}
\figcaption{Overview of existing MTMCT datasets.}{The table shows a comparison of Synthehicle to related datasets. A \cmark indicates whether 3D boxes, depth annotations and segmentations are included in the ground truth, respectively. For multi-camera datasets, the number of cameras corresponds to the total number of videos in the dataset. Naturally, single-camera datasets have one camera per scene, and their camera numbers are marked with *.}
\label{tab:data_comp}
\end{table*}

%% file: content/03_dataset.tex
\input{content/03_1_analysis}

\input{content/03_3_recording}

\input{content/03_2_data_types}

%% file: content/03_1_analysis.tex
\section{Dataset Analysis}

\subsection{Overview and Comparison}
So far, the MTMC vehicle tracking community has focused on \textit{tracking-by-detection}, a methodology based on 2D object detections and subsequent appearance feature extraction~\cite{qian2020electricity, luna2021online, tran2022robust, specker2022improving, yao2022city, li2022multi, yang2022box}. Unlike other datasets, which are designed for tracking targets enclosed by 2D bounding boxes, Synthehicle provides 3D annotations in terms of 3D bounding boxes and world coordinates, segmentations (instance, semantic and panoptic), and depth images. In Table~\ref{tab:data_comp}, we compare Synthehicle to established tracking datasets, such as PETS09, WILDTRACK and CityFlow. Synthehicle is the only dataset with all of the listed annotation types, and the longest dataset in duration. Compared to the closest related real dataset, CityFlow, Synthehicle is three times denser in terms of average number of vehicles per frame, has more than ten times as many scenes and fourteen times as many annotated bounding boxes. It also surpasses all other multi-camera tracking datasets in the number of cameras. Table~\ref{tab:overview} provides a detailed list of all Synthehicle scenes and their classification into train and test split.

\input{content/tables/scene_overview}

%% file: content/tables/scene_overview.tex
\begin{table*}[!htb]
    \begin{minipage}{.48\linewidth}
      \centering
        \resizebox{\linewidth}{!}{
\begin{tabular}{@{}lcrrrcc@{}}
\toprule
\textbf{Scene} & \textbf{\# Cams} & \textbf{\# Boxes} & \textbf{\# Tracks} & \textbf{Density} & \textbf{Test}\\ \midrule
    Town01-O-dawn    &     4 &   48\,048 &      82 &   6.67 & \xmark \\
    Town01-O-day     &     4 &   57\,539 &      80 &   7.99 & \xmark \\
    Town01-O-night   &     4 &   51\,725 &      73 &   7.18 & \xmark \\
    Town01-O-rain    &     4 &   70\,595 &      69 &   9.80 & \xmark \\
    Town02-O-dawn    &     3 &   61\,337 &      69 &  11.35 & \xmark \\
    Town02-O-day     &     3 &   64\,093 &      74 &  11.86 & \xmark \\
    Town02-O-night   &     3 &   47\,619 &      64 &   8.81 & \xmark \\
    Town02-O-rain    &     3 &   43\,416 &      72 &   8.04 & \xmark \\
    Town03-O-dawn    &     8 &  200\,088 &     126 &  13.89 & \xmark \\
    Town03-O-day     &     8 &  192\,994 &     124 &  13.40 & \xmark \\
    Town03-O-night   &     8 &  207\,035 &     117 &  14.37 & \xmark \\
    Town03-O-rain    &     8 &  192\,072 &     115 &  13.33 & \xmark \\
    Town04-O-dawn    &     4 &   33\,644 &      39 &   4.67 & \xmark \\
    Town04-O-day     &     4 &   35\,816 &      47 &   4.97 & \xmark \\
    Town04-O-night   &     4 &   28\,173 &      44 &   3.91 & \xmark \\
    Town04-O-rain    &     4 &   23\,086 &      42 &   3.20 & \xmark \\
    Town05-O-dawn    &     6 &  13\,1720 &      96 &  12.19 & \xmark \\
    Town05-O-day     &     6 &  11\,3413 &      96 &  10.50 & \xmark \\
    Town05-O-night   &     6 &  10\,7870 &      88 &   9.98 & \xmark \\
    Town05-O-rain    &     6 &  12\,7385 &      96 &  11.79 & \xmark \\
    Town06-O-dawn    &     4 &   19\,734 &      45 &   2.74 & \cmark \\
    Town06-O-day     &     4 &   13\,918 &      55 &   1.93 & \cmark \\
    Town06-O-night   &     4 &   15\,859 &      45 &   2.20 & \cmark \\
    Town06-O-rain    &     4 &   19\,308 &      57 &   2.68 & \cmark \\
    Town07-O-dawn    &     4 &   54\,679 &      58 &   7.59 & \cmark \\
    Town07-O-day     &     4 &   56\,314 &      57 &   7.82 & \cmark \\
    Town07-O-night   &     4 &   80\,797 &      46 &  11.22 & \cmark \\
    Town07-O-rain    &     4 &   46\,584 &      59 &   6.47 & \cmark \\
    Town10HD-O-dawn  &     5 &   95\,426 &     110 &  10.60 & \cmark \\
    Town10HD-O-day   &     5 &  16\,5259 &     116 &  18.36 & \cmark \\
    Town10HD-O-night &     5 &   89\,170 &     100 &   9.90 & \cmark \\
    Town10HD-O-rain  &     5 &   98\,855 &     125 &  10.98 & \cmark \\
\bottomrule
\end{tabular}}
    \end{minipage}%
    \hfill\begin{minipage}{.48\linewidth}
      \centering
        \resizebox{\linewidth}{!}{
\begin{tabular}{@{}lcrrrcc@{}}
\toprule
\textbf{Scene} & \textbf{\# Cams} & \textbf{\# Boxes} & \textbf{\# Tracks} & \textbf{Density}  & \textbf{Test} \\ \midrule
Town01-N-dawn    &     6 &   85\,515 &     143 &   7.91 & \xmark \\
Town01-N-day     &     6 &   77\,372 &     140 &   7.16 & \xmark \\
Town01-N-night   &     6 &   72\,420 &     140 &   6.70 & \xmark \\
Town01-N-rain    &     6 &   61\,648 &     130 &   5.70 & \xmark \\
Town02-N-dawn    &     5 &   50\,676 &      78 &   5.63 & \xmark \\
Town02-N-day     &     5 &   60\,452 &      83 &   6.71 & \xmark \\
Town02-N-night   &     5 &   59\,928 &      86 &   6.65 & \xmark \\
Town02-N-rain    &     5 &   56\,781 &      95 &   6.30 & \xmark \\
Town03-N-dawn    &     5 &   92\,879 &     182 &  10.31 & \xmark \\
Town03-N-day     &     5 &   79\,180 &     172 &   8.79 & \xmark \\
Town03-N-night   &     5 &   70\,648 &     166 &   7.84 & \xmark \\
Town03-N-rain    &     5 &   67\,393 &     157 &   7.48 & \xmark \\
Town04-N-dawn    &     5 &   53\,505 &     149 &   5.94 & \xmark \\
Town04-N-day     &     5 &   48\,521 &     134 &   5.39 & \xmark \\
Town04-N-night   &     5 &   52\,177 &     150 &   5.79 & \xmark \\
Town04-N-rain    &     5 &   56\,749 &     161 &   6.30 & \xmark \\
Town05-N-dawn    &     5 &   59\,804 &     131 &   6.64 & \xmark \\
Town05-N-day     &     5 &   61\,488 &     130 &   6.83 & \xmark \\
Town05-N-night   &     5 &   65\,224 &     142 &   7.24 & \xmark \\
Town05-N-rain    &     5 &   56\,459 &     122 &   6.27 & \xmark \\
Town06-N-dawn    &     7 &   41\,687 &     186 &   3.30 & \cmark \\
Town06-N-day     &     7 &   39\,087 &     188 &   3.10 & \cmark \\
Town06-N-night   &     7 &   37\,006 &     189 &   2.93 & \cmark \\
Town06-N-rain    &     7 &   44\,945 &     184 &   3.56 & \cmark \\
Town07-N-dawn    &     7 &   13\,936 &      42 &   1.10 & \cmark \\
Town07-N-day     &     7 &   17\,585 &      43 &   1.39 & \cmark \\
Town07-N-night   &     7 &   21\,916 &      56 &   1.73 & \cmark \\
Town07-N-rain    &     7 &   12\,354 &      41 &   0.98 & \cmark \\
Town10HD-N-dawn  &     7 &  134\,134 &     150 &  10.64 & \cmark \\
Town10HD-N-day   &     7 &  122\,552 &     149 &   9.72 & \cmark \\
Town10HD-N-night &     7 &  136\,116 &     137 &  10.80 & \cmark \\
Town10HD-N-rain  &     7 &  119\,476 &     144 &   9.48 & \cmark \\

\bottomrule
\end{tabular}}
\end{minipage}\\[0.1cm]
\figcaption{Overview over all 64 Synthehicle scenes.}{Scenes have a frame rate of 10fps, a resolution of 1920x1080 and a duration of 1800 frames per camera. The dataset is split into train and test scenes. The markers -O- and -N- in the scene names indicate overlapping and non-overlapping camera topology, respectively.}
\label{tab:overview}
\end{table*}

%% file: content/03_3_recording.tex
\subsection{Data Recording} CARLA~\cite{Dosovitskiy17} is an open-source simulation tool for building urban traffic scenarios and has been successfully employed to generate numerous synthetic datasets for computer vision tasks~\cite{DBLP:conf/cvpr/Varol0MMBLS17, DBLP:journals/corr/HandaPBSC15a, DBLP:conf/cvpr/RosSMVL16, DBLP:conf/cvpr/HuCHHS19, DBLP:conf/eccv/RichterVRK16, DBLP:conf/cvpr/Krahenbuhl18, DBLP:conf/cvpr/MarinVGL10, DBLP:conf/iciap/AmatoCFGM19, fabbri2018learning, kohl2020mta, fabbri2021motsynth, DBLP:journals/corr/GaidonWCV16, deschaud2021kitticarla, DBLP:conf/iccv/HuCWLSKDY19}. We utilize CARLA's rich set of realistic simulated sensors to render urban vehicle tracking scenarios using RGB, depth, and semantic LIDAR sensors in CARLA's eight pre-designed town maps. The data recording process can be described as follows.

First, we define two scenes for each of the eight towns - one for an overlapping camera view setup (O) and the other for a non-overlapping setup (N). We place a varying number of cameras (3 to 8) in each scene. The cameras are placed such that the vehicles are viewed from a high angle to mimic real-world positions, e.g., on top of traffic lights, similarly to CityFlow~\cite{tang2019cityflow}.
After defining the camera networks, we randomly spawn vehicles and pedestrians. The maximum number of vehicles that could spawn on a map is set to 200. Pedestrians mainly fulfill the task of influencing the otherwise monotonous flow of traffic. Models and appearances for vehicles and pedestrians are randomly chosen from CARLA's model pool and a list of realistic vehicle colors, matching real-life urban scenarios (cf. Figure~\ref{fig:carlacolors}). 

Vehicle routing and rules are controlled CARLA's \texttt{TrafficManager}, which was designed to manage vehicles in autopilot mode. Some vehicles will disobey traffic rules by driving too fast or crossing red traffic lights, providing more diverse and less predictable trajectories.
For every defined camera, we deploy an RGB sensor to record the traffic scenes with a resolution of $1920 \times 1080$ px and a frame rate of 10 fps. Semantic rotating LIDAR sensors capture information about all vehicles in a camera's field of view.  The sensor is implemented using ray-casting and exposes all information about objects hit by a ray. We use the sensor to filter out annotations for heavily occluded objects, e.g., for cars hidden behind buildings.
We attach two different segmentation sensors and a depth sensor to each RGB camera to record semantic and instance segmentations and to obtain depth information. The recording of all sensors lasts 1800 frames. 
Scenes are recorded as described above in four different ambience configurations: Day, dawn, rain, and night. Figure~\ref{fig:weatherman} shows an example frame recorded from the same camera under different configurations. When recording a scene for a weather configuration, all vehicle and pedestrian spawns and traffic flows will be randomized - only the sensor placement is constant for each scene. This way, Synthehicle provides a vast variety of traffic flows and vehicle trajectories. 
Finally, we extract perfect ground truth annotations while recording the scenes.

%% file: content/03_2_data_types.tex
\subsection{Data Types} 
\label{ss:datatypes}
We use CARLA's Python API to extract a variety of ground truth annotations: Camera calibrations, 2D and 3D detections, semantic, instance and panoptic segmentations, depth information and multi-camera tracking ground truth. 
\paragraph{Calibrations} For each camera, we obtain its $3 \times 3$ camera intrinsic matrix $\mathbf K$ and its $4 \times 4$ world-to-camera matrix $\bm M_{\text{w2c}} = [ \bm R, \bm t]$, which is the inverse of the camera extrinsic matrix. These matrices can be used for converting world points to image points and vice versa. We also provide $(x, y, z)$ 3D world positions for all cameras and their pitch, roll and yaw. 

\paragraph{3D Detections} For each object in the scene, we obtain its 3D bounding box in world $(x, y, z)$-coordinates directly from CARLA. The oriented bounding box is defined by eight corner points and yaw rotation. Each world point $(x_{\text w}, y_{\text w}, z_{\text w})^T \in \mathbb R^3$ is projected to camera coordinates using the $4 \times 4$ world-to-camera matrix $\bm M_{\text{w2c}}$ to obtain the point in image coordinates $(x, y)^T$ via 
\begin{align}
    \begin{bmatrix}
        \tilde{x}_{\text I} \\
        \tilde{y}_{\text I} \\
        \tilde{z}_{\text I} \\ 
        1
    \end{bmatrix} = \bm M_{\text{w2c}} \begin{bmatrix}
        x_{\text w} \\
        y_{\text w} \\
        z_{\text w} \\ 
        1
    \end{bmatrix}, && \begin{bmatrix}
        x_{\text I} \\
        y_{\text I} \\
        z_{\text I} 
    \end{bmatrix} = \bm K \begin{bmatrix}
        \tilde{y}_{\text I} \\
        - \tilde{z}_{\text I} \\ 
        \tilde{x}_{\text I}
    \end{bmatrix},
\end{align}
where $\bm K \in \mathbb R^{3\times 3}$ denotes the camera intrinsic matrix and $(x, y)^T = (x_I / z_I, y_I / z_I)^T$. Note that CARLA uses UnrealEngine's left-handed $z$-up coordinate system, which is why the axes have to be permuted and the $z$-axis reversed before multiplying with $\bm K$ to obtain $(x, y)^T$ in image space. We store 3D boxes in COCO-format~\cite{lin2014microsoft} as $\{ (x_i, y_i, z_i, w_i, h_i, l_i, \theta_i) \}_i$, where $(x_i, y_i, z_i)$ is the 3D center point, $(w_i, h_i, l_i)$ describe weight, height and length of the box, and $\theta_i$ is the yaw rotation.

\begin{figure}%
    \centering
    \subfloat[\centering 3D Bounding Box (2D projection)]{{\includegraphics[width=0.32\linewidth]{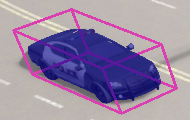} }}%
    \hfill
    \subfloat[\centering 2D Bounding Box (loose fit)]{{\includegraphics[width=0.32\linewidth]{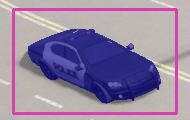} }}%
    \hfill
    \subfloat[\centering 2D Bounding Box (slim fit)]{{\includegraphics[width=0.32\linewidth]{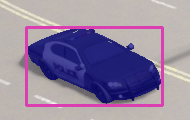} }}%
    \figcaption{Bounding Boxes in Synthehicle.}{The initial 3D bounding boxes (a) are transformed into 2D boxes (b) by a min-max procedure. Using semantic segmentation, this box is then refined to wrap tightly around the target object.}%
    \label{fig:boxes}%
\end{figure}

\paragraph{2D Detections} The 2D detections are obtained in image coordinates from the projected 3D detections. If $\mathcal B_{\text{3D}} = \{ \bm c_1, \dots, \bm c_8 \}$ is the 3D bounding box with its eight corner points, we obtain $(x_1, y_1, x_2, y_2)^T = (\min_x B_{\text{3D}}, \min_y B_{\text{3D}}, \max_x B_{\text{3D}}, \max_y B_{\text{3D}})^T$, i.e., by choosing the 2D box as the smallest rectangle containing all projected 3D vertices. Since 2D boxes acquired from this min-max-projection are usually larger than the enclosed target object, we tighten the box using the objects semantic segmentation label (see Figure~\ref{fig:boxes}). Bounding boxes smaller than $32 \times 32$ are filtered out from 2D detections, but are still kept in our general annotations.

\paragraph{Segmentations} CARLA provides two sensors for segmentation: A semantic segmentation and an instance segmentation camera. We capture images from both cameras for each frame and directly store the corresponding output images. Having obtained semantic and instance segmentation labels, the panoptic pixel labels can easily be calculated by looking up the corresponding segmentation pixel for each instance segmentation pixel. Figure~\ref{fig:example} illustrates an example of semantic and instance segmentation annotations.

\paragraph{Depth Buffer} Leveraging CARLA's depth sensor we also capture the depth map for each frame. The depth information can be used to improve camera projections since it provides scaling in depth dimension. Figure~\ref{fig:example} includes an example. 

\paragraph{Tracking} We store the tracking ground truth for 2D and 3D boxes in COCO and MOTChallenge~\cite{MOT16} format. Unlike CityFlow~\cite{tang2019cityflow}, our ground truth contains boxes for both vehicles and pedestrians, and boxes are included also if they are only visible in one camera. 

\begin{figure}
    \centering
    \includegraphics[width=0.9\linewidth]{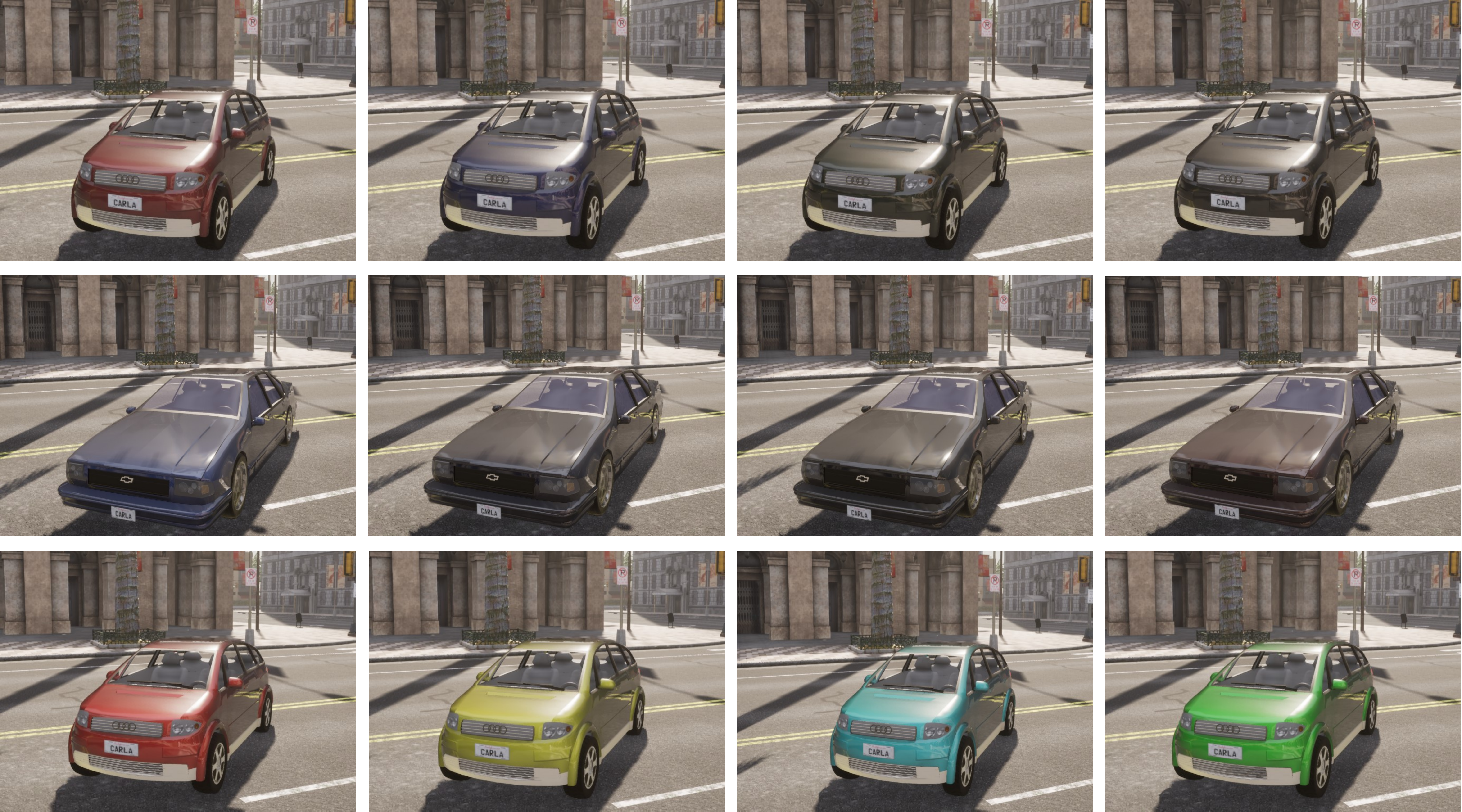}
    \figcaption{Vehicle colors in CARLA.}{The top two rows show vehicles from CARLA in the realistic color palette. The last row shows vehicles in a customized color palette. To increase difficulty of re-identification tasks, we only use realistic colors in Synthehicle.}%
    \label{fig:carlacolors}
\end{figure}

%% file: content/04_experiments.tex
\section{Experiments}

We conduct the following experiments to show the performance of existing methods on the Synthehicle dataset and to provide baselines for future research:  2D vehicle detection (Section \ref{ss:detection2d}), vehicle re-identification (Section \ref{ss:reid}), single-camera multi-vehicle tracking (Section \ref{ss:sct}), and  multi-vehicle multi-camera tracking (Subsection \ref{ss:mtmct}). 

\input{content/04_01_2D_Detection}
\input{content/04_03_Vehicle_ReID}
\input{content/04_04_SCT}
\input{content/04_05_MTMCT}

%% file: content/04_01_2D_Detection.tex
\subsection{2D Detection}
\label{ss:detection2d}
\input{content/tables/detection_results_2d}

An object detector processes an image $\bm I \in \mathbb R^{W \times H \times 3}$ and returns are set of distinct objects $\{ (x_i, y_i, w_i, h_i, c_i, s_i) \}_i$ with localization coordinates $(x_i, y_i, w_i, h_i)$, optionally with class $c_i$ and confidence score $s_i$. As such, object detectors are an important part of most tracking pipelines, which extract features from detected objects to match them accordingly. For smart cities, 2D vehicle detection is important for vehicle localization. 

\paragraph{Experimental Setup} We use the YOLOX-x model~\cite{yolox2021} in the mmdetection~\cite{mmdetection} framework. Research on training detectors on synthetic data suggest that variety is more important than data size~\cite{fabbri2021motsynth}. Therefore, for training and testing, we sample every tenth frame from all of our respective scenes and include it in the corresponding detection split.

\paragraph{Results} Table~\ref{tab:2ddet} lists the results of our detection training. When using pretrained COCO weights only and thereby not training on Synthehicle at all, we obtain mediocre detection results. By visualizing false positives and negatives, we observe that YOLOX-x struggles to detect special CARLA vehicles, such as fire engines and cybertrucks. However, when fine-tuning on the Synthehicle train splits, performance increases significantly. As expected, performance on rain and night scenes is worse than on day and dawn scenes. Interestingly, we obtain the highest performance on dawn scenes. This might be because in dawn scenarios there are almost no shadows present in the scenes, potentially reducing the number of false positives. Compared to training and testing on weather-specific train and test splits, training on all splits and then testing on the individual test splits leads to increased performance. Thus, the variety within the training data is crucial for a well-generalizing model.

%% file: content/tables/detection_results_2d.tex
\begin{table}[]
    \centering
    \resizebox{\linewidth}{!}{
    \begin{tabular}{llrrrrrr}
    \toprule
    \textbf{Train} & \textbf{Test} & \textbf{AP} & \textbf{AP}$_{50}$ & \textbf{AP}$_{75}$ & \textbf{AP}$_{S}$ & \textbf{AP}$_{M}$ & \textbf{AP}$_{L}$ \\
    \midrule
    -- & All & 0.242 & 0.438 & 0.217 & 0.003 & 0.149 & 0.460\\
    All & All & 0.597 & 0.842 & 0.651  & 0.151 & 0.480 & 0.785 \\
    \midrule
    Day & Day & 0.587 & 0.842 & 0.652 & 0.082 & 0.469 & 0.769 \\
    Dawn & Dawn & 0.608 & 0.866 & 0.660 & 0.139 & 0.505 & 0.792 \\
    Rain & Rain & 0.568 & 0.822 & 0.611 & 0.122 & 0.447 & 0.791 \\
    Night & Night & 0.506 & 0.780 & 0.540 & 0.119 & 0.381 & 0.666 \\
    \midrule
    All & Day & 0.626 & 0.870 & 0.693 & 0.116 & 0.510 & 0.808\\
    All & Dawn & 0.640 & 0.882 & 0.701 & 0.182 & 0.533 & 0.827\\
    All & Rain & 0.597 & 0.840 & 0.648 & 0.162 & 0.476 & 0.818\\
    All & Night & 0.522 & 0.777 & 0.560 & 0.171 & 0.383 & 0.696\\
    \bottomrule
    \end{tabular}}
    \figcaption{2D detection performance.}{We evaluate the YOLOX-x object detector with pretrained COCO weights under different train-test-split configurations. Values are given in \%.}
    \label{tab:2ddet}
\end{table}

%% file: content/04_03_Vehicle_ReID.tex
\subsection{Vehicle Re-Identification}
\label{ss:reid}
Object re-identification is an image retrieval problem in which a gallery set of $m$ images $\mathcal G = \{ \mathbf I_{G, i} \}_{i=1}^m$, $\mathbf I_{G, i} \in \mathbb R^{W \times H \times 3}$, is ranked by similarity to a query image $\mathbf I_Q \in \mathbb R^{W \times H \times 3}$, with the goal that the most similar gallery image belongs to the same identity (i.e., class) as the query image. In particular, images in object re-identification are obtained from multiple distinct cameras. The desired similarity measure in question is usually obtained by extracting features using a deep neural network and measuring their distances~\cite{huynh2021strong, zheng2019vehiclenet, Naphade20AIC20, Naphade21AIC21, DBLP:conf/icip/HerzogJTHGR21}. 

\input{content/tables/dataset_comp_reid}

Person and vehicle re-identification have received much attention in the past. Both tasks have to deal with similar problems like changing camera angles and lighting conditions. However, while people look and dress somewhat differently, vehicles can appear similarly. Additionally, vehicles provide almost no spatial-temporal pose information since, unlike humans, their shapes remain almost unchanged under movement. Important real datasets for vehicle re-identification are the VeRi dataset~\cite{liu2016large} with 776 vehicles captured over 20 cameras, and the AICity ReID dataset~\cite{tang2019cityflow, Naphade19AIC19}, which is provided as part of the AICityChallenge. 

\paragraph{Experimental Setup} We train vehicle re-identification on different train and test splits of Synthehicle using the fastreid~\cite{he2020fastreid} framework with a ResNet50 backbone and instance normalization~\cite{pan2018two}. Train and test splits were created by sampling every tenth frame and cropping detections using ground truth annotations. Crops smaller than $50 \times 50$ were filtered out. The vehicle images are scaled to $256\times 256$ pixels for the training and deformed if their bounding box is not quadratic. We always train up to 140 epochs with early stopping using the Adam optimizer~\cite{kingma2014adam}. During testing, we normalize the features in the Euclidean norm and use the cosine distance between feature vectors to calculate the cumulative matching characteristics (CMC)~\cite{moon2001computational}. We report the CMC rank-1 accuracy (r1) and mean Average Precision (mAP) for evaluation. As in training, the images are scaled to $256\times 256$ for testing.

\paragraph{Results} Table~\ref{tab:reid_results} lists the results produced by the fastreid model with different train and test combinations. As expected, re-identification works best in the day and dawn scenes and performs slightly worse in the rain and night scenes. If trained on all scenes, performance is increased on all test splits. This is surprising since CARLA only provides a limited number of vehicle models, and many spawned objects will be identical in appearance, but different in vehicle ID. The results indicate that the ``more data is better''-assumption of deep learning also holds in this particular case.

\input{content/tables/reid_results}

%% file: content/tables/dataset_comp_reid.tex
\begin{table}[!ht]
\centering
\resizebox{\linewidth}{!}{
\begin{tabular}{lrrrr}
\toprule
\textbf{Dataset} & \textbf{Training} & \textbf{Gallery} & \textbf{Query} & \textbf{\# Classes} \\ \midrule 
Synthehicle (day) & 24\,509 & 11\,556 & 2\,889 & 223 \\
Synthehicle (dawn) & 23\,963 & 8\,039 & 2\,010 & 220 \\
Synthehicle (rain) & 22\,171 & 7\,036 & 1\,760 & 205\\
Synthehicle (night) & 22\,943 & 8\,775 & 2\,194 & 205 \\
Synthehicle (all) & 93\,586 & 35\,407 & 8\,852 & 853 \\
\bottomrule
\end{tabular}}
\figcaption{Synthehicle Re-Identification Splits.}{We create five different splits for re-identification training.}
\label{tab:overview_reid}
\end{table}

%% file: content/tables/reid_results.tex
\begin{table}[]
\centering
\resizebox{\linewidth}{!}{
\begin{tabular}{@{}llrrrr@{}}
\toprule
\textbf{Train} & \textbf{Test} & \textbf{mAP (\%)} & \textbf{rank 1 (\%)} & \textbf{rank 5 (\%)} & \textbf{rank 10 (\%)} \\ \midrule
All & All & 47.82 & 51.55 & 72.29 & 80.03 \\
\midrule
Day & Day & 59.89 & 62.02 & 77.12 & 85.04\\
Dawn & Dawn & 47.57 & 51.17 & 72.30 & 80.47 \\
Rain & Rain & 39.08 & 48.44 & 72.31 & 80.29\\
Night & Night & 27.04 & 36.84 & 58.76 & 67.41 \\
\midrule 
All & Day &  60.38  & 58.50 & 78.56 & 84.58 \\
All & Dawn & 53.21 & 55.75 & 82.51 & 53.21\\
All & Rain & 45.33 & 52.60 & 75.66 & 83.76\\
All & Night & 32.69 & 37.25 & 60.78 & 68.88\\
\bottomrule
\end{tabular}}\\[0.2cm]
\figcaption{Vehicle Re-Identification Performance}{
We used different splits of training, fine tuning and testing.}
\label{tab:reid_results}
\end{table}

%% file: content/04_04_SCT.tex
\subsection{Single-Camera 2D Tracking}
\label{ss:sct}
In single-camera multi-object tracking-by-detection, the task is to generate consistent tracks from a time-ordered set of images $\mathcal I = \{ \bm I_t \}_t$ and corresponding object detections $\mathcal D = \{ D_t \}_t$, $D_t = \{(x_i, y_i, w_i, h_i)_i\}_t$ (cf. Section \ref{ss:detection2d}), such that every distinct target is assigned an unambiguous and unique ID across all time frames. Occlusions, lighting variations, and false positive or negative detections make tracking challenging. Important trackers include, among others, DeepSORT~\cite{wojke2017deepsort}, Tracktor \cite{tracktor_2019_ICCV}, and CenterTrack \cite{zhou2020tracking}. 

\paragraph{Experimental Setup} We use the DeepSORT~\cite{wojke2017deepsort} tracker with the YOLOX-x models trained in Section~\ref{ss:detection2d} and the re-identification weights obtained from the experiments in Section~\ref{ss:reid}. We follow the evaluation protocols of single-camera tracking by utilizing the CLEAR MOT metrics~\cite{bernardin2008mota}. We set the minimum detection height and width to $32 \times 32$ to filter small false positive detections and set the detection confidence threshold to $0.3$. The DeepSORT parameter \texttt{nn\_budget}, which decides how many appearance features are considered for a track, is set to $100$.

\paragraph{Results} Table~\ref{tab:sct} shows the results of DeepSORT on all test scenes. Note that results are averaged over the respective number of camera videos included in a scene. As expected, tracking in day and dawn scenes generally yields superior performance than tracking on rain and night scenes. However, for Town06-O-night, DeepSORT performs better than in other scenes with the same camera setup. We conjecture that the 10 fps frame rate is sufficiently large to track objects in a scene with comparatively low density (cf. Table~\ref{tab:overview}). Note also that scenes with different ambience configurations are rendered with different vehicle spawns and number of vehicles each. Only the camera setup is fixed. Overall, DeepSORT-based single-camera tracking on Synthehicle generates satisfying baseline results.

\input{content/tables/single_camera_tracking_results}

%% file: content/tables/single_camera_tracking_results.tex
\begin{table*}[!ht]
\centering
\resizebox{\linewidth}{!}{
\begin{tabular}{lrrrrrrrrrrrrrr}
\toprule
 & \multicolumn{10}{c}{\textbf{Single-Camera}} & \multicolumn{3}{c}{\textbf{Multi-Camera}} \\
\cmidrule(lr){2-11} 
\cmidrule(lr){12-14}
             Scene &  MOTA &   MOTP &   IDF1 &    IDP &    IDR &   GT &   MT &   PT &   ML &   IDs & IDF1 & IDP & IDR \\
\midrule
    Town06-N-day &  54.3\% &  0.175 &  65.8\% &  78.4\% &  56.8\% &  438 &  116 &  194 &  128 &   252  & 48.2\% & 58.1\% & 41.6\%\\
    Town06-N-dawn &  55.0\% &  0.204 &  64.4\% &  80.4\% &  53.7\% &  437 &  126 &  176 &  135 &   335  & 49.3\% & 61.8\% & 41.2\%\\
    Town06-N-rain &  50.5\% &  0.191 &  60.4\% &  77.2\% &  49.6\% &  474 &  124 &  204 &  146 &   389  & 41.4\% & 53.9\% & 34.0\%\\
   Town06-N-night &  45.8\% &  0.183 &  59.9\% &  76.2\% &  49.4\% &  455 &  105 &  187 &  163 &   257  & 39.6\% & 51.1\% & 32.7\%\\
   \midrule
     Town07-N-day &  62.0\% &  0.157 &  78.1\% &  79.3\% &  76.9\% &  142 &   77 &   59 &    6 &    35 & 46.6\% & 47.3\% & 45.9\%\\
    Town07-N-dawn &  73.4\% &  0.173 &  80.6\% &  90.3\% &  72.8\% &  141 &   68 &   66 &    7 &    46 & 59.2\% & 66.3\% & 53.4\% \\
    Town07-N-rain &  62.3\% &  0.158 &  78.4\% &  79.6\% &  77.2\% &  122 &   55 &   64 &    3 &    45 & 57.8\% & 59.2\% &  57.0\%\\
   Town07-N-night &  50.6\% &  0.198 &  62.4\% &  83.9\% &  49.7\% &  195 &   51 &  109 &   35 &   154 & 40.8\% & 54.8\% & 32.5\%\\
   \midrule
   Town10HD-N-day &  53.8\% &  0.186 &  65.9\% &  68.8\% &  63.2\% &  566 &  219 &  255 &   92 &   846 & 38.8\% & 40.9\% & 37.2\%\\
  Town10HD-N-dawn &  54.3\% &  0.204 &  63.7\% &  67.5\% &  60.2\% &  579 &  221 &  274 &   84 &  1157 & 36.3\% & 39.0\% & 34.3\%\\
  Town10HD-N-rain &  50.8\% &  0.201 &  63.7\% &  66.2\% &  61.5\% &  565 &  192 &  277 &   96 &   954 & 37.7\% & 40.5\% & 36.4\%\\
 Town10HD-N-night &  54.8\% &  0.192 &  68.6\% &  69.2\% &  68.0\% &  355 &  140 &  150 &   65 &  1004 & 50.8\% & 55.3\% & 50.4\%\\
 \midrule
     Town06-O-day &  68.5\% &  0.145 &  72.2\% &  85.5\% &  62.5\% &  150 &   47 &   50 &   53 &    35 & 46.7\% & 55.3\% & 40.4\%\\
Town06-O-dawn &  66.4\% &  0.147 &  72.2\% &  87.4\% &  61.5\% &  143 &   53 &   58 &   32 &    53 & 45.4\% & 55.0\% & 38.7\%\\
    Town06-O-rain &  58.6\% &  0.153 &  68.3\% &  86.6\% &  56.3\% &  172 &   48 &   75 &   49 &    61 & 45.9\% & 60.5\% & 37.8\%\\
   Town06-O-night &  69.7\% &  0.142 &  72.4\% &  84.8\% &  63.2\% &  125 &   45 &   44 &   36 &    55 & 50.1\% & 58.8\% & 43.7\%\\
   \midrule
     Town07-O-day &  71.5\% &  0.149 &  72.7\% &  80.3\% &  66.3\% &  171 &   92 &   54 &   25 &   257 & 37.1\% & 41.0\% & 33.8\%\\
    Town07-O-dawn &  74.7\% &  0.145 &  77.4\% &  83.7\% &  72.0\% &  190 &  117 &   56 &   17 &   212 & 39.6\% & 43.4\% & 36.8\%\\
    Town07-O-rain &  65.6\% &  0.156 &  66.7\% &  75.2\% &  59.9\% &  172 &   77 &   67 &   28 &   256 & 42.6\% & 49.1\% & 38.3\%\\
   Town07-O-night &  50.4\% &  0.213 &  58.3\% &  62.7\% &  54.5\% &  127 &   56 &   46 &   25 &  1194 & 30.3\% & 32.9\% & 28.3\%\\
   \midrule
   Town10HD-O-day &  61.9\% &  0.197 &  71.2\% &  74.7\% &  68.0\% &  437 &  205 &  159 &   73 &   702 & 25.8\% & 27.1\% & 24.6\%\\
  Town10HD-O-dawn &  64.9\% &  0.201 &  68.0\% &  72.6\% &  63.9\% &  454 &  226 &  184 &   44 &   665 & 31.9\% & 34.9\% & 30.0\%\\
  Town10HD-O-rain &  47.3\% &  0.212 &  61.2\% &  68.7\% &  55.1\% &  519 &  194 &  239 &   86 &   740 & 30.9\% & 36.1\% & 27.8\%\\
 Town10HD-O-night &  50.2\% &  0.220 &  58.1\% &  61.5\% &  55.1\% &  431 &  163 &  230 &   38 &   908 & 24.5\% & 30.7\% & 23.3\%\\
\bottomrule
\end{tabular}}
\figcaption{Single-Camera and Multi-Camera Tracking Performance.}{Performance of DeepSORT on single cameras of the respective scenes and ELECTRICITY on multiple cameras. Single-camera results are obtained for every individual camera in the scene, and then averaged for this table.}
\label{tab:sct}
\end{table*}

%% file: content/04_05_MTMCT.tex
\subsection{Multi-Target Multi-Camera 2D Tracking}
\label{ss:mtmct}

MTMCT is the extension of single-camera tracking to a multiple-camera setup, where the input set of images now comes from $K$ distinct views $\mathcal I = (\{ \bm I_t^{(c_1)} \}_t, \dots \{ \bm I_t^{(c_K)} \}_t)$, and the detections are obtained accordingly as $\mathcal D = \{\mathcal D^{(c_i)}\}_{i=1,\dots,K}$. Generally, there is no restriction on the camera topology and field of views can be overlapping or non-overlapping. In practice, many MTMC trackers~\cite{DBLP:conf/cvpr/RistaniT18, qian2020electricity} first apply a single-camera tracker on the individual camera videos to obtain single-camera results (i.e., tracklets) $\mathcal T_1, \dots, \mathcal T_K$, and then match those sets of local tracklets into a set of global tracklets $\mathcal T$.

\paragraph{Method} We use a method based on ELECTRICITY~\cite{qian2020electricity}, a 2020 challenge winner for the CityFlow dataset, with the single-camera tracklets $\mathcal T_1, \dots, \mathcal T_K$ obtained by applying DeepSORT as in Section~\ref{ss:sct}. For each tracklet, a query image is chosen, and all other images of that tracklet are selected to be gallery images. Using the feature extractor trained in Section~\ref{ss:reid}, a query feature matrix $\bm Q \in \mathbb R^{n \times f}$ and a gallery feature matrix $\bm G \in \mathbb R^{m \times f}$ are built, where $n$ is the number of tracklets, $m$ is the number of gallery images, and $f$ is the feature embedding size. After normalizing all query and probe features, \ie $\| \bm q_i \|_2 = 1 ~ \forall i=1, \dots, n$ and $\| \bm g_j \|_2 = 1 ~\forall j=1, \dots, m$, we obtain the cosine distance matrix $\bm D$ as
\begin{align}
    \bm D = \bm Q \bm G^T \in \mathbb R^{n \times m}.
\end{align}

The MVMCT result is then derived from $\mathbf D$ by merging the tracklets $\mathcal T_i$ and $\mathcal T_j$ if and only if both the feature distance between the query image of $\mathcal T_i$ and the gallery images of $\mathcal T_j$, and the feature distance between the query image of $\mathcal T_j$ and the gallery images of $T_i$ are below a specified threshold $\theta$, \ie,
\begin{equation}
    \bm D_{ij} < \theta \text{~and~} \bm D_{ji} < \theta.
\end{equation}

\paragraph{Experimental Setup} We apply the method described above with $\theta=0.8$ on all test scenes. For evaluation, we report the IDF1, IDP and IDR metrics for multi-camera tracking as suggested by~\cite{ristani2016performance}.

\begin{figure}%
    \centering
    \subfloat[\centering ]{{\includegraphics[width=0.48\linewidth]{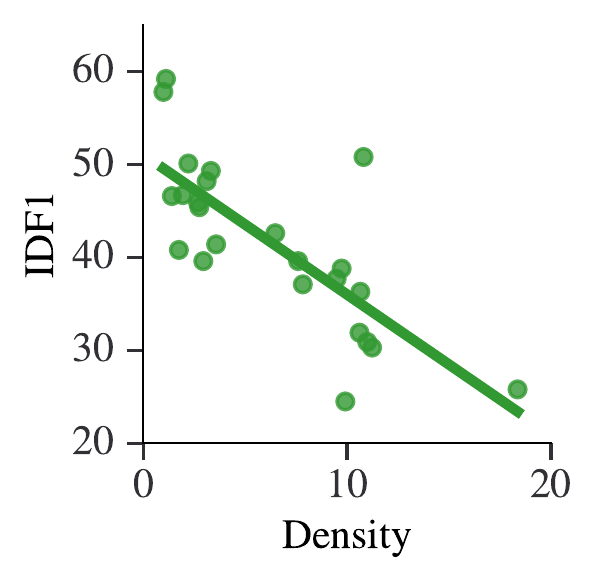}}}%
    \hfill
    \subfloat[\centering ]{{\includegraphics[width=0.48\linewidth]{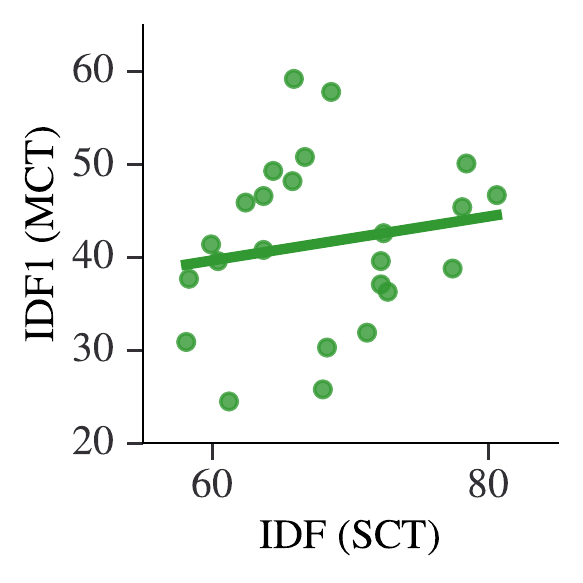} }}%
    \figcaption{Results analysis.}{The scene density has a strong influence on multi-camera tracking performance (a). The quality of single-camera tracklets influences the multi-camera tracking performance, but less than expected (b).}%
    \label{fig:density}%
\end{figure}

\paragraph{Results} Table~\ref{tab:sct} lists the performance of the multi-camera tracker on Synthehicle. Since the tracker is based on pre-extracted single-camera tracklets, there is a correlation between single-camera performance and multi-camera performance. The multi-camera tracker relies solely on re-identification and performance sometimes decreases in night scenes (e.g., in Town10HD-O-night). Figure~\ref{fig:density} shows the influence of scene density and single-camera IDF1 performance on multi-camera IDF1 performance. Dense scenes increase the difficulty for our multi-camera tracking, since more gallery and query features have to be considered during matching, leading to potential association errors.

%% file: content/05_conclusion.tex
\section{Conclusion} We have presented a massive synthetic dataset for tracking vehicles across multiple overlapping and non-overlapping cameras in various scenes with a wide variety of data annotations previously not included in similar datasets, such as semantic, instance and panoptic segmentations, depth maps, and over 4 million annotated 2D and 3D bounding boxes. With 17 hours of video material and 340 cameras, it is the largest available multi-target multi-camera tracking dataset. The ambiance configurations included in our dataset allow for exploring multi-vehicle tracking under challenging conditions. We have demonstrated the performance of different baselines for vehicle detection, re-identification, and single- and multi-camera tracking tasks. Results on these tasks indicate that Synthehicle is a complex dataset with diverse and challenging scenarios.
While the focus of our analysis was multi-target multi-camera tracking, our annotations can potentially enable the exploration of new tasks, such as 3D multi-target multi-camera tracking or multi-camera tracking and segmentation.

\section{Acknowledgement} We gratefully acknowledge the financial support from Deutsche Forschungsgemeinschaft (DFG) under grant number RI 658/25-2.